%% file: main.tex
\documentclass[]{scrartcl}
\usepackage[left=1in, right=1in, top=1in, bottom=1in]{geometry}
\usepackage[dvipsnames]{xcolor}
\usepackage{microtype}

\usepackage[english]{babel}
\usepackage{csquotes}
\usepackage{authblk}

\usepackage{booktabs} 

\usepackage{bm}
\usepackage{dirtytalk}
\usepackage{amsfonts}
\usepackage{physics}
\usepackage[overload]{textcase}
\usepackage{algorithm}
\usepackage{algpseudocode}
\usepackage[normalem]{ulem}
\usepackage[subrefformat=parens,labelformat=parens]{subcaption}

\usepackage{tikz} 
\usetikzlibrary{arrows.meta}

\usepackage{pgfplots}
\usepackage{pgfplotstable}
\usepgfplotslibrary{fillbetween}
\usepgfplotslibrary{ternary}
\pgfplotsset{compat=1.16}
\usepgfplotslibrary{external}
\usepackage[colorlinks=true, citecolor=Blue, linkcolor=BrickRed, urlcolor=BrickRed]{hyperref}
\usepackage[capitalize]{cleveref}

\usepackage[style=authoryear, natbib=true, sorting=nyt, maxbibnames=9, maxcitenames=2, uniquelist=false, backend=biber]{biblatex}
\title{Probabilistic task modelling for meta-learning}
\subtitle{}
\author[1]{Cuong Nguyen}
\author[2]{Thanh-Toan Do}
\author[1]{Gustavo Carneiro}
\affil[1]{%
    Australian Institute for Machine Learning, University of Adelaide, Australia
}
\affil[2]{%
    Department of Data Science and AI, Monash University, Australia
}
\date{}

\begin{document}
    \maketitle

    \input{Abstract.tex}
    \input{Introduction.tex}
    \input{RelatedWork}
    \input{Methodology}
    \input{Experiments}
    \input{Conclusion}
    \section*{Acknowledgement}
        This work was supported by Australian Research Council through grants DP180103232 and FT190100525. We also acknowledge the Phoenix HPC service at the University of Adelaide that provided the super-computing resources for this work.

    \printbibliography
    
    \newpage
    \onecolumn
    \input{Appendices}
\end{document}

%% file: Abstract.tex
\begin{abstract}
    We propose \emph{probabilistic task modelling} -- a generative probabilistic model for collections of tasks used in meta-learning. The proposed model combines variational auto-encoding and latent Dirichlet allocation to model each task as a mixture of Gaussian distribution in an embedding space. Such modelling provides an explicit representation of a task through its task-theme mixture. We present an efficient approximation inference technique based on variational inference method for empirical Bayes parameter estimation. We perform empirical evaluations to validate the \emph{task uncertainty} and \emph{task distance} produced by the proposed method through correlation diagrams of the prediction accuracy on testing tasks. We also carry out experiments of task selection in meta-learning to demonstrate how the task relatedness inferred from the proposed model help to facilitate meta-learning algorithms.
\end{abstract}

%% file: Introduction.tex
\section{Introduction}
\label{sec:introduction}
    The latest developments in machine learning have enabled the field to solve increasingly complex classification problems. 
    Such complexity require high capacity models, which in turn need a massive amount of annotated data for training, resulting in an arduous, costly and even infeasible annotation process. 
    This has, therefore, motivated the research of novel learning approaches, generally known as transfer learning, that exploit past experience (in the form of models learned from other training tasks) to quickly learn a new task using relatively small training sets.
    
\begin{figure}[t]
        \centering
        \begin{tikzpicture}
            \begin{axis}[
                height=0.3 \linewidth,
                width=0.3 \linewidth,
                ybar,
                ymin=0,
                align=center,
                xtick={0.1, 0.3, 0.5, 0.7},
                ytick={0, 1, 2, 3, 4},
                xlabel={Accuracy},
                ylabel={Normalised frequency},
                scale only axis,
            ]
                \addplot [hist={density, bins=10},
                fill=orange!75,
                draw=orange!50!black]
                table [y index=0] {results/accuracy.txt};
            \end{axis}
        \end{tikzpicture}
        \caption{The results locally produced for MAML on 15,504 available 5-way 1-shot mini-ImageNet testing tasks vary from 20 to 70 percent accuracy, suggesting that not all testing tasks are equally related to training tasks.}
        \label{fig:histogram_maml}
    \end{figure}
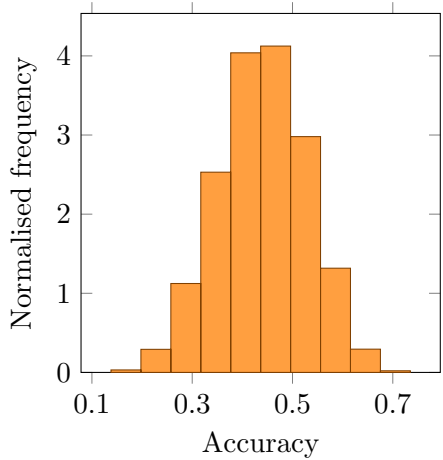

    Transfer-learning, and in particular, meta-learning, assumes the existent of a task environment where training and testing tasks are i.i.d. sampled from the same latent distribution. By modelling such environment through meta-parameters that are shared across all tasks, meta-learning can solve an unseen task more accurately and efficiently by learning how to solve many tasks generated from the same distribution, even if each task contains a limited number of training examples. Meta-learning has, therefore, progressed steadily with many remarkable state-of-the-art results in several few-shot learning benchmarks~\citep{vinyals2016matching,snell2017prototypical,finn2017model,yoon2018bayesian,rusu2019meta, allen2019infinite}. However, current development of meta-learning focuses on solving tasks without providing understanding on how tasks are generated or correlated, potentially leading to sub-optimal solutions. In fact, there is a large variation of prediction performance made by various meta-learning algorithms reported in~\citep[\figureautorefname~1]{dhillon2019baseline} or shown in \figureautorefname~\ref{fig:histogram_maml}, suggesting that not all testing tasks are equally related to training tasks. This motivates our work to model and represent tasks in a latent \say{task-theme} space. The new task representation allows further downstream works, such as task similarity or active task selection, to be developed to gain insights into, or even improve, the prediction performance of different meta-learning algorithms.
    
    In this paper, we propose \textbf{probabilistic task modelling} (PTM) -- a graphical model that combines variational auto-encoding (VAE)~\citep{kingma2013auto} and Gaussian latent Dirichlet allocation (LDA)~\citep{das2015gaussian} -- to \textbf{model tasks} used in meta-learning\footnote{This is a comprehensive work extended from our workshop paper~\citep{nguyen2021similarity}}. Note that PTM itself is not a meta-learning method. With this modelling approach, the dataset associated with each task can be modelled as a mixture of finite Gaussian distributions, allowing to represent tasks in a latent \say{task-theme} simplex via its mixture coefficient vector. Such representation provides a convenient way to quantitatively measure \say{task uncertainty} or task relatedness, which can be utilised in active task selection for meta-learning.

%% file: RelatedWork.tex
\section{Related Work}
\label{sec:related_work}
    The proposed approach is closely related to Task2Vec \citep{achille2019task2vec} when modelling tasks for meta-learning. In Task2Vec, a task is represented by an embedding computed from the Fisher information matrix associated with the task-specific classifier. In PTM, a task is represented by the variational distribution of task-theme mixture, which is a part of the graphical model describing the task generation process. The two methods, therefore, differ at the modelling mechanism: Task2Vec follows a deterministic approach, while PTM is a probabilistic method. Such difference provides an advantage of PTM over Task2Vec, which includes modelling uncertainty into the task representation. In addition, PTM is more efficient than Task2Vec at inference when predicting task representation, since PTM only needs a single forward pass, while Task2Vec requires to re-train or fine-tune the task-specific classifier and calculate the Fisher information matrix for the task that needs to be presented. There is also a slightly similar work recently published about task modelling~\citep{kaddour2020probabilistic}.
    
    Our work is related to task similarity estimation, which has been intensively studied in the field of multi-task learning. Some remarkable examples in this area include task-clustering using k-nearest neighbours~\citep{thrun1996discovering}, or modelling common prior between tasks as a mixture of distributions~\citep{bakker2003task,xue2007multi}. Another approach is to formulate  multi-task learning as a convex optimisation problem either to cluster tasks and utilise the clustering results to fast track the learning~\citep{jacob2009clustered}, or to learn task relationship through task covariance matrices~\citep{zhang2012convex}. Other approaches provided theoretical guarantees when learning the similarity or relationship between tasks~\citep{shui2019a}. Recently, the \emph{taskonomy} project~\citep{zamir2018taskonomy} was conducted to carry out extensive experiments on 26 computer-vision tasks to empirically analyse the correlation between those tasks. Other works~\citep{tran2019transferability,nguyen2020leep} take a slightly different approach by investigating the correlation of the label distributions between the tasks of interest to measure task similarity. One commonality among all studies above is their reliance on task-specific classifiers which are used to quantify task relatedness. In contrast, our proposed method explicitly models tasks without the help of any task-specific classifier, making it more efficient in training and prediction.
    
    Our work is also connected to finite mixture models~\citep{pritchard2000inference}, such as the \emph{latent Dirichlet allocation}~\citep{blei2003latent}, which analyses and summarises text data in topic modelling, or categorises natural scenes in computer vision~\citep{li2005bayesian}. LDA assumes that each document within a given corpus can be represented as a mixture of finite categorical distributions, where each categorical distribution is a latent topic shared across all documents. Training an LDA model or its variants on a large text corpus is challenging, so several approximate inference techniques have been proposed, ranging from mean-field variational inference (VI)~\citep{blei2003latent}, collapsed Gibbs' sampling~\citep{griffiths2004finding} and collapsed VI~\citep{teh2007collapsed}. Furthermore, several online inference methods have been developed to increase the training efficiency for large corpora~\citep{canini2009online,hoffman2010online,foulds2013stochastic}. Our work is slightly different from the modelling of the conventional LDA, where we do not use the data directly, but embed it into a latent space. In short, our approach is a combination of VAE~\citep{kingma2013auto} and LDA to model the dataset associated with a task. Our approach considers \say{word} as continuous data, instead of the discrete data represented by a bag-of-word vector generally used by LDA-based topic modelling methods. The resultant model in the embedding latent space is, therefore, similar to the Gaussian LDA~\citep{das2015gaussian} for word embedding in topic modelling.

%% file: Methodology.tex
\section{Probabilistic task modelling}
\label{sec:methodology}
\begin{figure}[t]
        \centering
        \begin{tikzpicture}
            \begin{ternaryaxis}[
                height = 0.5 \linewidth,
                width = 0.5 \linewidth,
                colorbar horizontal,
                colorbar style = {height = 0.75em},
                colormap name = viridis,
                ternary limits relative=false,
                xmin=0, xmax=1, xlabel = {Furniture},
                ymin=0, ymax=1, ylabel = {Animal},
                zmin=0, zmax=1, zlabel = {Human},
                label style={sloped},
                table/col sep=comma, table/header=true,
                point meta rel=per plot,
                ]
                
                \addplot3[patch, patch type=triangle, shader=interp, point meta=\thisrow{pdf}] table {img/ternary_mesh.csv};
                
                \node [circle, minimum size=4pt, inner sep=0pt, draw=White, fill=Red, pin={[pin edge={very thick}]110:Task \(i\)}] at (0.4, 0.125) {};
            \end{ternaryaxis}
        \end{tikzpicture}
        \caption{An example of a task-theme simplex where each task is represented by a 3-dimensional mixture vector.}
        \label{fig:task_theme_simplex}
    \end{figure}
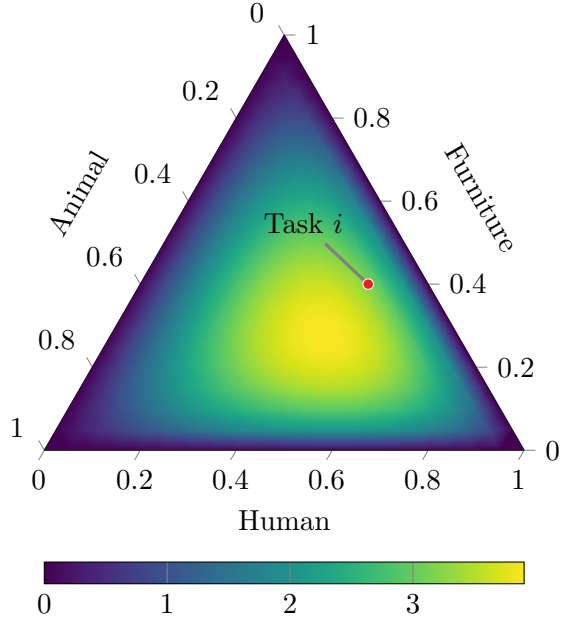

    To relate \textbf{task modelling} to \textbf{topic modelling}, we consider \textbf{a task} as \textbf{a document}, and \textbf{a data point} as \textbf{a word}. Given these analogies, we can use LDA~\citep{blei2003latent} -- a popular topic model -- to model tasks for meta-learning. However, simply applying LDA for task modelling would not scale well with high-dimensional data and large datasets. We, therefore, propose to employ the VAE~\citep{kingma2013auto} to reduce the dimension of the data, and use the inferred embeddings of data as words to model tasks. Due to the nature of VAE, the latent variables are often continuous, not discrete as the bag-of-words used in the conventional LDA. We, therefore, \textbf{replace} the \emph{categorical} \textbf{word-topic} distributions \textbf{in LDA} by \emph{Gaussian} \textbf{task-theme}\footnote{\say{Task-theme} is inspired by \citet{li2005bayesian}} distributions. Given these assumptions, each task can be modelled as a mixture of \(K\) Gaussian task-themes, allowing to represent tasks by their inferred \emph{task-theme mixture} vectors in the latent task-theme simplex as illustrated in \figureautorefname~\ref{fig:task_theme_simplex}. Hence, it is beneficial to utilise this representation for further downstream tasks, such as measuring task similarity.
    
\begin{figure}[t]
        \centering
        \begin{tikzpicture}[state/.style={circle, draw=black, minimum size=1cm}, scale=1, every node/.style={scale=1}]
    		\input{img/meta_gmm_vae}
    	\end{tikzpicture}
        \caption{The graphical model used in task modelling. The solid arrows denote data generation, while the dashed arrows stand for inference. The boxes are \protect\say{plates} representing replicates. The shading nodes denote observable variables, while the white nodes denote latent variables.}
        \label{fig:meta_gmm_vae}
    \end{figure}
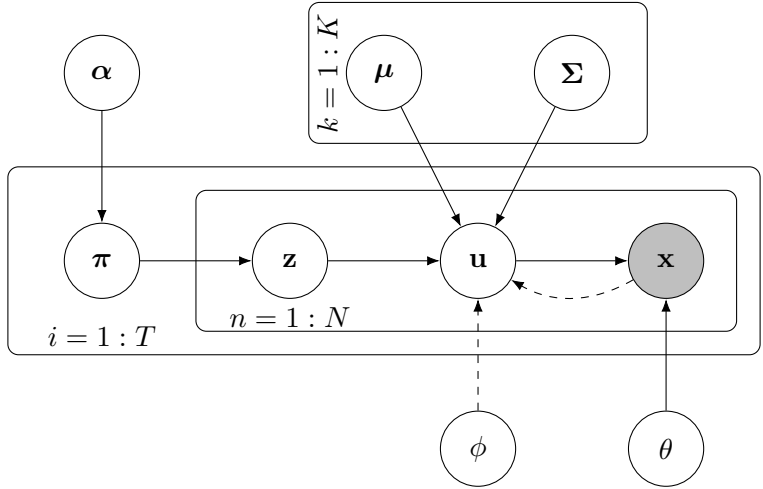

    The graphical model of the proposed PTM is shown in \figureautorefname~\ref{fig:meta_gmm_vae}, where there are \(T\) tasks, and each task consists of \(N\) data points, denoted as \(\mathbf{x}\). To simplify the formulation and analysis, \(N\) is assumed to be fixed across all tasks, but the extension of varying \(N\) is trivial and can be implemented straightforwardly. Under these assumptions, a task can be generated as follows:
	\begin{itemize}
	    \item Initialise the Dirichlet prior for task-theme mixture: \(\{\alpha_{k}\}_{k=1}^{K}\), where \(\alpha \in \mathbb{R}_{+}\)
        \item Initialise means and covariance matrices of \(K\) Gaussian task-themes \(\{\bm{\mu}_{k}, \bm{\Sigma}_{k}\}_{k=1}^{K}\), where \(\bm{\mu}_{k} \in \mathbb{R}^{D}\),  \(\bm{\Sigma}_{k} \in \mathbb{R}^{D \times D}\) is positive definite matrix, and \(D\) is the dimension of the data embedding
		\item For task \(\mathcal{T}_{i}\) in the collection of \(T\) tasks:
		\begin{itemize}
			\item Choose a task-theme mixture vector: \(\bm{\pi}_{i}~\sim~\mathrm{Dirichlet} \left( \bm{\pi}; \bm{\alpha}\right) \)
			\item For data point \(n\)-th of task \(\mathcal{T}_{i}\):
			\begin{itemize}
				\item Choose an task-theme assignment one-hot vector: \(\mathbf{z}_{in}~\sim~\mathrm{Categorical} \left( \mathbf{z}; \bm{\pi}_{i} \right)\)
				\item Choose an embedding of the data point: \({\mathbf{u}_{in} \sim \mathcal{N}\left(\mathbf{u}; \bm{\mu}_{k}, \bm{\Sigma}_{k}\right)} \), where: \(z_{ink} = 1\)
				\item Generate the data point from a decoder \(h\) parameterised by \(\theta\): \({\mathbf{x}_{in} = h(\mathbf{u}_{in}; \theta)}\).
			\end{itemize}
		\end{itemize}
	\end{itemize}

	To model tasks according to the task generation described above, we need to infer the task-agnostic (or meta) parameters of interest, namely \(\bm{\mu}, \bm{\Sigma}, \bm{\alpha}\) and  \(\theta\). However, due to the complexity of the graphical model shown in \figureautorefname~\ref{fig:meta_gmm_vae}, the exact inference for the posterior \(p(\bm{\mu}, \bm{\Sigma}, \bm{\alpha}, \theta | \mathbf{x})\) is intractable, and therefore, the estimation must rely on approximate inference. For simplicity, maximum likelihood estimation (MLE) is used as the objective function:
	\begin{equation}
	    \max_{\bm{\mu}, \bm{\Sigma}, \bm{\alpha}, \theta} \ln p(\mathbf{x} | \bm{\mu}, \bm{\Sigma}, \bm{\alpha}, \theta).
	    \label{eq:mle}
	\end{equation}
	
	Although MLE simplifies the learning for the meta-parameters of interest, the log-likelihood in~\eqref{eq:mle} is still difficult to evaluate for the optimisation. One workaround solution is to find its lower-bound, and maximise the lower-bound instead of maximising the log-likelihood itself. This approach is analogous to the variational inference, which has been widely used to infer the latent parameters of VAE and LDA models.

	Since the proposed PTM is a combination of VAE and LDA, the derivation for the lower-bound of the likelihood in \eqref{eq:mle} can be divided into 2 steps, where the first step is analogous to the lower bound of a VAE, and the second step is similar to the plain LDA model.
	
	In the first step, the latent variable \(\mathbf{u}\) is introduced, so that the log-likelihood \(\ln p(\mathbf{x} | \bm{\mu}, \bm{\Sigma}, \bm{\alpha}, \theta)\) can be bounded below by Jensen's inequality:
	\begin{equation}
	    \ln p(\mathbf{x} | \bm{\mu}, \bm{\Sigma}, \bm{\alpha}, \theta) \ge \mathsf{L}_{\mathrm{VAE}},
	\end{equation}
	where the lower-bound is defined as:
	\begin{equation}
	    \begin{aligned}[b]
	        \mathsf{L}_{\mathrm{VAE}} = \mathbb{E}_{q(\mathbf{u})} \left[ \ln p \left( \mathbf{x} | \mathbf{u}, \theta \right) + \ln p \left( \mathbf{u} | \bm{\mu}, \bm{\Sigma}, \bm{\alpha} \right) - \ln q(\mathbf{u}) \right],
	    \end{aligned}
	    \label{eq:vae_elbo}
	\end{equation}
	with \(q(\mathbf{u})\) being the variational distribution for the latent variable \(\mathbf{u}\).
	
	Following the conventional VI for VAE~\citep{kingma2013auto}, the variational distribution for the latent embedding \(\mathbf{u}\) is assumed to be a Gaussian distribution with diagonal covariance matrix:
	\begin{equation}
	    q(\mathbf{u}) = \mathcal{N} \left(\mathbf{u}; \mathbf{m}, \mathrm{diag}\left(\mathbf{s}^{2}\right) \right) \label{eq:q_u}.
	\end{equation}
	
	In addition, the parameters \(\mathbf{m}\) and \(\mathbf{s}\), which represent the distribution encoding the data \(\mathbf{x}\), are modelled by a neural network (also known as an encoder) \(f\) parameterised by \(\phi\):
	\begin{equation}
	    \begin{bmatrix}\mathbf{m}^{\top} & \mathbf{s}^{\top} \end{bmatrix}^{\top} = f(\mathbf{x}; \phi).
	\end{equation}
	
	Hence, instead of maximising the marginal log-likelihood in \eqref{eq:mle}, the lower-bound in \eqref{eq:vae_elbo} is maximised, resulted in the alternative objective:
	\begin{equation}
	    \max_{\bm{\mu}, \bm{\Sigma}, \bm{\alpha}, \theta} \max_{\phi} \mathsf{L}_{\mathrm{VAE}}.
	    \label{eq:objective_vae_elbo}
	\end{equation}
	
	One difficulty in maximising the lower-bound in \eqref{eq:objective_vae_elbo} is the evaluation for the prior \(\ln p \left( \mathbf{u} | \bm{\mu}, \bm{\Sigma}, \bm{\alpha} \right)\) of the latent embedding \(\mathbf{u}\) shown in Eq.~\eqref{eq:vae_elbo}. In vanilla VAE, the embedding prior is often modelled as some standard distributions, such as Gaussian or Beta, resulting in a tractable solution. In this paper, the prior is modelled as a Gaussian mixture model, making the solution intractable. However, since this prior is the marginal log-likelihood in the conventional LDA model, we can apply techniques developed for LDA methods to approximate this term. Here, we employ the VI approach in which the term is bounded below by Jensen's inequality:
	\begin{equation}
        \ln p \left( \mathbf{u} | \bm{\mu}, \bm{\Sigma}, \bm{\alpha} \right) \ge \mathsf{L}_{\mathrm{LDA}}(\mathbf{u}, q(\mathbf{z}, \bm{\pi})),
	\end{equation}
	where:
	\begin{equation}
	    \begin{aligned}[b]
	        & \mathsf{L}_{\mathrm{LDA}} \left( \mathbf{u}, q(\mathbf{z}, \bm{\pi}) \right) = \mathbb{E}_{q(\mathbf{z}, \bm{\pi})} \left[ \ln p(\mathbf{u} | \mathbf{z}, \bm{\mu}, \bm{\Sigma})  + \ln p(\mathbf{z} | \bm{\pi}) + \ln p(\bm{\pi} | \bm{\alpha}) - \ln q(\mathbf{z}) - \ln q(\bm{\pi}) \right],
	    \end{aligned}
	    \label{eq:lda_elbo}
	\end{equation}
	with \(q(\mathbf{z}, \bm{\pi})\) being the variational distribution for \(\mathbf{z}\) and \(\bm{\pi}\). This corresponds to the second step in the derivation.
	
	Similar to LDA~\citep{blei2003latent}, the variational distribution \(q(\mathbf{z}, \bm{\pi})\) is assumed to be fully factorised and followed the conjugate priors:
	\begin{equation}
	    q(\mathbf{z}, \bm{\pi}) = \prod_{i=1}^{T} q(\bm{\pi}_{i}; \bm{\gamma}_{i}) \prod_{n=1}^{N} q(\mathbf{z}_{in}; \mathbf{r}_{in}),
	\end{equation}
	where:
	\begin{align}
	    q(\bm{\pi}_{i}; \bm{\gamma}_{i}) & = \mathrm{Dirichlet}\left( \bm{\pi}_{i}; \bm{\gamma}_{i} \right) \label{eq:q_pi}\\
	    q(\mathbf{z}_{in}; \mathbf{r}_{in}) & = \mathrm{Categorical} \left( \mathbf{z}_{in}; \mathbf{r}_{in} \right) \label{eq:q_z},
	\end{align}
	with \(\mathbf{r}\) and \(\bm{\gamma}\) being the parameters of the variational distribution \(q(\mathbf{z}, \bm{\pi})\).
	
	In practice, \(q(\mathbf{z}, \bm{\pi})\) is obtained as the maximiser of the lower-bound \(\mathsf{L}_{\mathrm{LDA}} \left( \mathbf{u}, q(\mathbf{z}, \bm{\pi}) \right)\) on the embedding data \(\mathbf{u}\). A naive implementation to infer such variational distribution is problematic since the data is used twice: one to optimise \(q(\mathbf{z}, \bm{\pi})\), and the other is to optimise for \(q(\mathbf{u})\) in \eqref{eq:objective_vae_elbo}, which may result in overfitting. To avoid this issue, we employ the \emph{empirical Bayes} approach by splitting the training dataset of each task into two halves, where one half of data in a task, denoted as \(\mathbf{u}^{(t)}\), is used to obtain \(q(\mathbf{z}, \bm{\pi})\), while the other half, denoted as \(\mathbf{u}^{(v)}\), is used for the optimisation in \eqref{eq:objective_vae_elbo}. This approach is analogous to the \emph{empirical Bayes} meta-learning~\citep{finn2017model,nguyen2020uncertainty}, where one part of data is used for task-adaptation (often known as \say{inner-loop}), while the other part is used to learn the meta-parameter (often known as \say{outer-loop}).
	
	Given this modelling approach, the objective function can be formally written as a bi-level optimisation:
	\begin{equation}
	    \begin{aligned}[b]
	        \max_{\bm{\mu}, \bm{\Sigma}, \bm{\alpha}, \theta, \phi} & \mathsf{L}\left(\mathbf{u}^{(v)}, q^{*}\left(\mathbf{z}, \bm{\pi}\right) \right)\\
	        \text{s.t.: } & q^{*}\left(\mathbf{z}, \bm{\pi}\right) = \arg\max_{q(\mathbf{z}, \bm{\pi})} \mathbb{E}_{q(\mathbf{u}^{(t)}; \phi)} \left[ \mathsf{L}_{\mathrm{LDA}} \left( \mathbf{u}^{(t)}, q(\mathbf{z}, \bm{\pi}) \right) \right],
	    \end{aligned}
	    \label{eq:objective_vi}
	\end{equation}
	where
	\begin{equation}
    	\begin{split}
    	    & \mathsf{L} \left(\mathbf{u}^{(v)}, q^{*}\left(\mathbf{z}, \bm{\pi}\right) \right) = \mathbb{E}_{q \left( \mathbf{u}^{(v)}; \phi \right)} \left[ \mathsf{L}_{\mathrm{LDA}} \left(\mathbf{u}^{(v)}, q^{*}(\mathbf{z}, \bm{\pi})\right) + \ln p \left( \mathbf{x}^{(v)} | \mathbf{u}^{(v)}, \theta \right) - \ln q \left(\mathbf{u}^{(v)}; \phi \right) \right].
	   \end{split}
	   \label{eq:upper_level_function}
	\end{equation}
	
	Due to the assumptions made in \cref{eq:q_u,eq:q_pi,eq:q_z}, prior conjugate can be applied to simplify the evaluation for all the terms in \eqref{eq:lda_elbo} w.r.t. the variational distribution \(q(.)\). Details of the evaluation can be referred to \appendixname{~\ref{sec:elbo_calculation}}. In addition, the optimisation for the meta-parameters in \eqref{eq:objective_vi} is based on gradient ascent, and carried out in two steps, resulting in a process analogous to the expectation-maximisation (EM) algorithm. In the E-step (corresponding to the optimisation for the lower-level in \eqref{eq:objective_vi}), the task-specific variational-parameters \(\mathbf{r}\) and \(\bm{\gamma}\) are iteratively updated, while holding the meta-parameters \(\bm{\mu}, \bm{\Sigma}, \bm{\alpha}, \theta\) and \(\phi\) fixed. In the M-step (corresponding to the optimisation for the upper-level), the meta-parameters are updated using the values of the task-specific variational-parameters obtained in the E-step. Note that the inference for the task-theme parameters \(\bm{\mu}\) and \(\bm{\Sigma}\) are similar to the estimation of Gaussian mixture model~\citep[Chapter 9]{bishop2006pattern}. Please refer to \appendixname{~\ref{sec:elbo_optimisation}} for more details on the optimisation.
    	
	Conventionally, the iterative updates in the E-step and M-step require a full pass through the entire collection of tasks. This is, however, very slow and even infeasible since the number of tasks, \(T\), is often in the magnitude of millions. We, therefore, propose an online VI inspired by the online learning for LDA~\citep{hoffman2010online} to make the solution more scalable when inferring the meta-parameters. For each task \(\mathcal{T}_{i}\), we perform the EM to obtain the \say{task-specific} parameters (denoted by a tilde on top of variables) that are locally optimal for that task. The \say{meta} parameters of interest are then updated as a weighted average between their previous values and the \say{task-specific} values:
	\begin{equation}
	    \begin{aligned}[b]
	        \bm{\mu} & \gets (1 - \rho_{i}) \bm{\mu} + \rho_{i} \Tilde{\bm{\mu}}\\
	        \bm{\Sigma} & \gets (1 - \rho_{i}) \bm{\Sigma} + \rho_{i} \Tilde{\bm{\Sigma}}\\
	        \bm{\alpha} & \gets \bm{\alpha} - \rho_{i} \underbrace{\Tilde{\bm{\alpha}}_{i}}_{\mathbf{H}^{-1} \mathbf{g}},
	    \end{aligned}
	    \label{eq:online_update}
	\end{equation}
	where \(\rho_{i} = (\tau_{0} + i)^{-\tau_{1}}\) with \(\tau_{0} \ge 0\) and \(\tau_{1} \in (0.5, 1]\) \citep{hoffman2010online}, \(\mathbf{g}\) is the gradient of \(\mathsf{L}_{\mathrm{LDA}}\) w.r.t. \(\bm{\alpha}\), and \(\mathbf{H}\) is the Hessian matrix. The learning for the encoder \(\phi\) and the decoder \(\theta\) follows the conventional learning by stochastic gradient ascent. The complete learning algorithm for the proposed probabilistic task modelling is shown in \cref{alg:online_ptm}.

    Also, instead of updating the meta-parameters as in \eqref{eq:online_update} when observing a single task, we use multiple or a mini-batch of tasks to reduce the effect of measurement noise. The mini-batch version requires a slight modification in the formulation presented above, where we calculate the average of all \say{task-specific} parameters for tasks in the same mini-batch, and use that as the \say{task-specific} value to update the corresponding \say{meta} parameters.
	
	\begin{algorithm}[t]
        \caption{Online probabilistic task modelling}
        \label{alg:online_ptm}
        \begin{algorithmic}[1]
            \Procedure{Training}{} 
            \State Initialise LDA parameters: \(\{\bm{\mu}_{k}, \bm{\Sigma}_{k}, \alpha_{k} \}_{k=1}^{K}\)
            \State Initialise encoder \(\phi\) and decoder \(\theta\)
            \For{each mini-batch of \(T_{\mathrm{mini}}\) tasks}
                \For{\(i=1:T_{\mathrm{mini}}\)}
                    \State Split data into \(\{\mathbf{x}_{i}^{(t)}, \mathbf{y}_{i}^{(t)}\}\) and \(\{\mathbf{x}_{i}^{(v)}, \mathbf{y}_{i}^{(v)}\}\)
                    \State \(\mathbf{m}^{(t)}_{i}, \mathbf{s}_{i}^{(t)} \gets f(\mathbf{x}_{i}^{(t)}; \phi)\) 
                    \State \(\mathbf{m}^{(v)}_{i}, \mathbf{s}_{i}^{(v)} \gets f(\mathbf{x}_{i}^{(v)}; \phi)\)
                    \State \(\bm{\gamma}, \mathbf{r} \gets\) \Call{E-step}{$\mathcal{N}(\mathbf{u}; \mathbf{m}_{i}^{(t)}, \mathrm{diag}((\mathbf{s}_{i}^{(t)})^{2}))$}
                    \State Calculate \(\mathsf{L}\left(\mathbf{u}_{i}^{(v)}, q^{*}_{i} \left(\mathbf{z}, \bm{\pi}\right) \right)\) \Comment{Eq.~\eqref{eq:upper_level_function}}
                    \State Calculate \say{local} task-themes \(\Tilde{\bm{\mu}}_{i}, \Tilde{\bm{\Sigma}}_{i}, \Tilde{\bm{\alpha}}_{i}\)
                \EndFor
                \State \(\overline{\mathsf{L}} = \frac{1}{T} \sum_{i=1}^{T} \mathsf{L}\left(\mathbf{u}_{i}^{(v)}, q^{*}_{i} \left(\mathbf{z}, \bm{\pi}\right) \right)\)
                \State \(\bm{\mu}, \bm{\Sigma}, \bm{\alpha} \gets \mathrm{online\_LDA}\left( \Tilde{\bm{\mu}}_{1:T}, \Tilde{\bm{\Sigma}}_{1:T}, \Tilde{\bm{\alpha}}_{1:T} \right)\) 
                \State \(\theta, \phi \gets \mathrm{SGD}\left( -\overline{\mathsf{L}} \right)\) \Comment{gradient ascent}
            \EndFor
            \State \textbf{return} \(\bm{\mu}, \bm{\Sigma}, \bm{\alpha}, \theta, \phi\)
            \EndProcedure
            
            \Statex
            
            \Procedure{E-step}{$\mathcal{N}(\mathbf{u}; \mathbf{m}, \mathrm{diag}(\mathbf{s}^{2}))$}
                \State Initialise \(\mathbf{r}, \bm{\gamma}\)
                \Repeat
                    \State calculate the un-normalised \(r_{ink}\) \Comment{Eq.~\eqref{eq:r}}
                    \State normalise \(\mathbf{r}_{in}\) such that \(\sum_{k=1}^{K} r_{ink} = 1\)
                    \State calculate \(\gamma_{ik}\) \Comment{Eq.~\eqref{eq:gamma}}
                \Until{\(\frac{1}{K}~| \text{change in } \bm{\gamma} | < \text{threshold}\) }
                \State \textbf{return} \(\bm{\gamma}, \mathbf{r}\)
            \EndProcedure
        \end{algorithmic}
    \end{algorithm}
    
    Although the \say{reconstruction} term \(\ln p(\mathbf{x}^{(v)} | \mathbf{u}^{(v)}, \theta)\) in \eqref{eq:objective_vi} is used to model the likelihood of un-labelled data, it can straightforwardly be extended to a labelled data pair \(\{\mathbf{x}^{(v)}, \mathbf{y}^{(v)}\}\) by introducing the parameter \(\mathbf{w}\) of a classifier. In that case, the \say{reconstruction} term can be expressed as:
	\begin{equation}
    	\begin{aligned}[b]
    	    \ln p(\mathbf{x}^{(v)}, \mathbf{y}^{(v)} | \mathbf{u}^{(v)}, \theta, \mathbf{w}) = \underbrace{\ln p(\mathbf{y}^{(v)} | \mathbf{u}^{(v)}, \mathbf{w})}_{\text{negative classification loss}} + \underbrace{\ln p(\mathbf{x}^{(v)} | \mathbf{u}^{(v)}, \theta)}_{\text{negative reconstruction loss}}.
    	\end{aligned}
	\end{equation}
	In general, \(\mathbf{w}\) can be either a task-specific parameter generated from an additional meta-parameter shared across all tasks -- corresponding to \emph{empirical Bayes} meta-learning (e.g. using train-test split to learn hyper-parameters) algorithms~\citep{finn2017model,nguyen2020uncertainty}, or a meta-parameter itself -- corresponding to \emph{metric} meta-learning~\citep{vinyals2016matching,snell2017prototypical}. For simplicity, we will use the latter approach relying on the prototypical network~\citep{snell2017prototypical} with Euclidean distance on the data embedding \(\mathbf{u}\), to calculate the classification loss on labelled data. This reduces the need to introduce an additional parameter \(\mathbf{w}\) into our modelling.
	
	\subsection*{Task representation}
	    Given the inferred meta-parameters, including the task-themes \(\{\bm{\mu}_{k}, \bm{\Sigma}_{k} \}_{k=1}^{K} \), the Dirichlet prior \(\{ \bm{\alpha}_{l} \}_{l=1}^{L}\), the encoder \(\phi\) and the decoder \(\theta\), we can embed the data of a task into a latent space, and calculate its variational Dirichlet posterior of the task-theme \emph{mixing coefficients} \(q(\bm{\pi}; \bm{\gamma}_{i})\). The obtained distribution can be used represent the corresponding task in the latent task-theme simplex as illustrated in \figureautorefname~\ref{fig:task_theme_simplex}. This new representation of tasks has two advantages comparing to the recently proposed task representation Task2Vec~\citep{achille2019task2vec}: (i) it explicitly models and represents tasks without the need of any pre-trained networks to use as a \say{probe} network, and (ii) it uses a probability distribution, instead of a vector as in Task2Vec, allowing to include modelling uncertainty when representing tasks. Given the probabilistic nature of PTM, we can use the entropy of the inferred task-theme mixture distribution \(q(\bm{\pi}; \bm{\gamma}_{i})\) as a measure of \emph{task uncertainty}. In \sectionautorefname~\ref{sec:correlation_diagram}, we empirically show that this measure correlates to the generalisation or test performance.
    
        In addition, the representation produced by PTM can be used to quantitatively analyse the similarity or \emph{distance} between two tasks \(i\) and \(j\) through a divergence between \(q(\bm{\pi}; \bm{\gamma}_{i})\) and \(q(\bm{\pi}; \bm{\gamma}_{j})\). Commonly, symmetric distances, such as Jensen-Shannon divergence, Hellinger distance, or earth's mover distance are employed to calculate the divergence between distributions. However, it is argued that similarity should be represented as an asymmetric measure~\citep{tversky1977features}. This is reasonable in the context of transfer learning, since knowledge gained from learning a difficult task might significantly facilitate the learning of an easy task, but the reverse might not always have the same level of effectiveness. In light of asymmetric distance, we decide to use Kullback-Leibler (KL) divergence, denoted as \(D_{\mathrm{KL}}[. \Vert .]\), to measure \emph{task distance}. As \(D_{\mathrm{KL}} \left[ P \Vert Q \right]\) is defined as the information lost when using a code optimised for \(Q\) to encode the samples of \(P\), we, therefore, calculate \(D_{\mathrm{KL}} \left[ q(\bm{\pi}; \bm{\gamma}_{T + 1}) \Vert  q(\bm{\pi}; \bm{\gamma}_{i}) \right]\), where \(i \in \{1, \ldots, T\}\), to assess how the training task \(\mathcal{T}_{i}\) differs from the learning of the novel task \(\mathcal{T}_{T + 1}\).

%% file: img/meta_gmm_vae.tex
\pgfmathsetmacro{\yshift}{2.5}
\pgfmathsetmacro{\xshift}{2.5}
\pgfmathsetmacro{\Rmin}{0.75}
\node[state] at (0, 0) (pi) {\(\bm{\pi}\)};
\node[state] at ([xshift=0, yshift=\yshift cm]pi) (alpha) {\(\bm{\alpha}\)};

\node[state] at ([xshift=\xshift cm, yshift=0cm]pi) (z) {\(\mathbf{z}\)};

\node[state] at ([xshift=\xshift cm, yshift=0cm]z) (u) {\(\mathbf{u}\)};

\node[state, fill=gray!50] at ([xshift=1 * \xshift cm, yshift=0cm]u) (x) {\(\mathbf{x}\)};

\node[state] at ([xshift=-0.5 * \xshift cm, yshift=\yshift cm]u) (mu) {\(\bm{\mu}\)};
\node[state] at ([xshift=0.5 * \xshift cm, yshift=\yshift cm]u) (Lambda) {\(\bm{\Sigma}\)};


\node[state] at ([xshift= 0. * \xshift cm, yshift=-\yshift cm]x) (theta) {\(\theta\)};

\node[state] at ([xshift=0 * \xshift cm, yshift=- \yshift cm]u) (phi) {\(\phi\)};

\draw[-Latex] (alpha) -- (pi);
\draw[-Latex] (pi) -- (z);
\draw[-Latex] (z) -- (u);
\draw[-Latex] (mu) -- (u);
\draw[-Latex] (Lambda) -- (u);


\draw[-Latex] (u) -- (x);
\draw[-Latex] (theta) -- (x);

\draw[-Latex, dashed] (phi) -- (u);
\draw[-Latex, dashed] (x) to [bend left] (u);

\draw[rounded corners] ([xshift=-0.5 * \xshift cm, yshift=-0.375 * \yshift cm]z) rectangle ([xshift=0.375 * \xshift cm, yshift=0.375 * \yshift cm]x);

\draw[rounded corners] ([xshift=-0.5 * \xshift cm, yshift= -0.5 * \yshift cm]pi) rectangle ([xshift=0.5 * \xshift cm, yshift=.5 * \yshift cm]x);

\draw[rounded corners] ([xshift=-.4 * \xshift cm, yshift=-0.375 * \yshift cm]mu) rectangle ([xshift=.4 * \xshift cm, yshift=.375 * \yshift cm]Lambda);

\node[align=left] at ([xshift=0cm, yshift=-0.3 * \yshift cm]z) {\(n=1:N\)};

\node[align=left] at ([xshift=0cm, yshift=-0.4 * \yshift cm]pi) {\(i=1:T\)};

\node[align=left, rotate=90] at ([xshift=-.3 * \xshift cm, yshift=0 * \yshift cm]mu) {\(k=1:K\)};

%% file: Experiments.tex
\section{Experiments}
\label{sec:experiments}
    In this section, we empirically validate the two properties of PTM -- \emph{task uncertainty} and \emph{task distance} -- through task distance matrix and correlation diagrams. We also show two applications of the proposed approach used in active task selection for inductive and transductive life-long meta-learning. The experiments are based on the \(n\)-way \(k\)-shot tasks formed from Omniglot~\citep{lake2015human} and mini-ImageNet~\citep{vinyals2016matching} -- the two widely used datasets  to evaluate the performance of  meta-learning algorithms.
    
    The Omniglot dataset consists of 1623 different handwritten characters from 50 different alphabets, where each character was drawn in black and white by 20 different people. Instead of using random train-test split that mixes all characters, the original split~\citep{lake2015human} is used to yield finer-grained classification tasks. In addition to the task forming based on randomly mixing characters of many alphabets, the two-level hierarchy of alphabets and characters are utilised to increase the difficulty of the character classification. Note that no data augmentations, such as rotating images by multiples of 90 degrees, is used throughout the experiments. Also, all images are down-sampled to 64-by-64 pixel\(^{2}\) to simplify the image reconstruct in the decoder.
    
    The mini-ImageNet dataset comprises a small version of ImageNet, which contains 100 classes taken from ImageNet, and each class has 600 colour images. We follow the common train-test split that uses 64 classes for training, 16 classes for validation, and 20 classes for testing~\citep{ravi2017optimization}. Similar to Omniglot, all images are also in 64-by-64 pixel\(^{2}\).
    
    The encoder used in the experiments consists of 4 convolutional modules, where each module has a convolutional layer with 4-by-4 filters and 2-by-2 stride, followed by a batch normalisation and a leaky rectified linear activation function with a slope of 0.01.  The output of the last convolutional layer is flattened and connected to a fully connected layer to output the desired dimension for the latent variable \(\mathbf{u}\). The decoder is designed similarly, except that the convolutional operator is replaced by the corresponding transposed convolution. For the Omniglot dataset, the number of filters within each convolutional layer of the encoder is 8, 16, 32, and 64, respectively, and the dimension of \(\mathbf{u}\) is 64. For mini-ImageNet dataset, these numbers are 32, 64, 128 and 256, and the dimension of \(\mathbf{u}\) is 128. The reconstruction loss follows the negative log-likelihood of the continuous Bernoulli distribution~\citep{loaiza2019continuous}, which is often known as binary cross-entropy, while the classification loss is based on the prototypical network used in metric learning. The training subset of each task, \(\mathbf{u}^{(t)}_{i}\), is used to calculate the class prototypes, and the classification loss is based on the soft-max function of the distances between the encoding of each input image to those prototypes~\citep{snell2017prototypical}. The optimiser used is Adam with the step size of \(2 \times 10^{-4}\) to optimise the parameters of the encoder and decoder after every mini-batch consisting of 20 tasks. For the LDA part, a total of \(K = 8\) task-themes is used. The Dirichlet prior is assumed to be symmetric with a concentration \(\alpha = 1.1\) across both datasets. The parameters of the learning rate used in the online LDA are \(\rho_{0} = 10^{6}\) and \(\rho_{1} = 0.5\). A total of \(10^{6}\) episodes are used to train PTM on both datasets. We note that setting \(\alpha > 1\) enforces every task to be modelled as a mixture of many task-themes, avoiding the task-themes collapsing into a single task-theme during training. The phenomenon of task-theme collapse when \(\alpha < 1\) is not observed in LDA, but in PTM due to the integration of VAE. At the beginning of training, the encoder is inadequate, producing mediocre embedding features. The resulting features, combined with \(\alpha < 1\), makes a task more likely to be represented by a single task-theme. By learning solely from that task-theme, the encoder is pushed to bias further toward to that task-theme, making only one task-theme distribution updated, while leaving others unchanged. When \(\alpha > 1\), all the task-themes contribute to the representation of a task, so they can be learnt along with the encoder. Please refer to \href{https://github.com/cnguyen10/probabilistic_task_modelling}{https://github.com/cnguyen10/probabilistic\_task\_modelling} for the implemented code.

    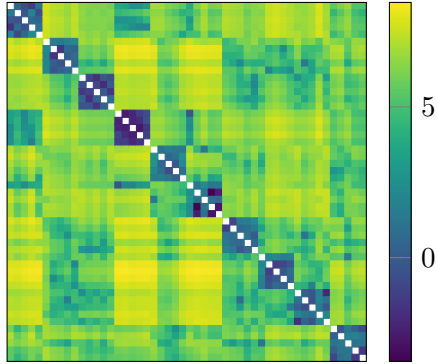
\begin{figure}[t]
        \centering
        \begin{tikzpicture}
            \pgfmathsetmacro{\ncols}{50}
            \begin{axis}[
                enlargelimits = false,
                height = 0.3 \linewidth,
                width = 0.3 \linewidth,
                ticks = none,
                colorbar,
                colorbar style = {width = 0.75em},
                colormap name = viridis,
                scale only axis,
                ymin = -0.5,
                ymax = \ncols - 0.5,
                xmin = -0.5,
                xmax = \ncols - 0.5
            ]
                \addplot [
                matrix plot,
                mesh/cols=\ncols,
                point meta=explicit,
                ] table [meta=distance, col sep=comma, header=true] {results/distance_matrix_mesh_50x50.csv};
            \end{axis}
        \end{tikzpicture}
        \caption{The matrix of log KL distances between Omniglot tasks shows that tasks that are generated from the same alphabet are closer together, denoted as the dark green blocks along the diagonal. The matrix is asymmetric due to the asymmetry of the KL divergence used as the task distance.}
        \label{fig:task_distance_matrix}
    \end{figure}

    \begin{figure*}[t]
            \hspace{-0.5em}
            \begin{subfigure}[b]{0.2 \linewidth}
                \begin{tikzpicture}
                    \tikzstyle{every node}=[font=\small]
                    \pgfmathsetmacro{\slope}{-0.01446709}
                    \pgfmathsetmacro{\intercept}{0.4905035577312847}
                    \pgfmathsetmacro{\stdScale}{1}
                    \pgfmathsetmacro{\newSlope}{100 * \slope}
                    \pgfmathsetmacro{\newIntercept}{100 * \intercept}
                    
                    \pgfplotstableread[col sep=comma, header=true]{results/Entropy_omniglot.csv} \myTable
                    
                    \begin{axis}[
                        height = \linewidth,
                        width = \linewidth,
                        clip marker paths=true,
                        title = {Omniglot},
                        title style = {yshift=-0.5em},
                        enlarge y limits=true,
                        xlabel = {Entropy of test tasks},
                        ylabel = {Prediction accuracy (\%)},
                        ylabel style = {yshift=-1em},
                        legend style={draw=none, anchor=north, at={(0.5, -0.3)}},
                        table/col sep=comma,
                        table/header=true,
                        scale only axis
                    ]
                        \addplot[only marks, color=MidnightBlue, mark size=2pt, forget plot] table [x={x}, y expr={\thisrow{accuracy} * 100}] {\myTable}; 

                        \addplot[sharp plot, color=BurntOrange, very thick] table [x={xtest}, y expr={\thisrow{ymean} * 100}] {\myTable}; 

                        \addplot[name path=std_upper, draw=none] table [x={xtest}, y expr=100 * (\thisrow{ymean} + \stdScale * \thisrow{ystd})] {\myTable}; 
                        \addplot[name path=std_lower, draw=none] table [x={xtest}, y expr=100 * (\thisrow{ymean} - \stdScale * \thisrow{ystd})] {\myTable}; 
                        \addplot [fill=BurntOrange, opacity=0.25] fill between [of=std_upper and std_lower];

                        \pgfmathsetmacro{\newSlope}{100 * \slope}
                        \pgfmathsetmacro{\newIntercept}{100 * \intercept}
                        \addlegendentry{\(y = \pgfmathprintnumber{\newSlope} \cdot x
                        \pgfmathprintnumber[print sign]{\newIntercept}\)}
                    \end{axis}
                \end{tikzpicture}
                \caption{}
                \label{fig:task_entropy_omniglot}
            \end{subfigure}
            \hspace{3em}
            \begin{subfigure}[b]{0.2 \linewidth}
                \begin{tikzpicture}
                    \tikzstyle{every node}=[font=\small]
                    \pgfmathsetmacro{\slope}{-0.04402093}
                    \pgfmathsetmacro{\intercept}{-0.3891100720048166}
                    \pgfmathsetmacro{\stdScale}{1}
                    \pgfmathsetmacro{\newSlope}{100 * \slope}
                    \pgfmathsetmacro{\newIntercept}{100 * \intercept}
                    
                    \pgfplotstableread[col sep=comma, header=true]{results/Entropy_miniImageNet_64.csv} \myTable

                    \begin{axis}[
                        height = \linewidth,
                        width = \linewidth,
                        clip marker paths=true,
                        title = {Mini-ImageNet},
                        title style = {yshift=-0.5em},
                        enlarge y limits=true,
                        xlabel = {Entropy of test tasks},
                        legend style={draw=none, anchor=north, at={(0.5, -0.3)}},
                        table/col sep=comma,
                        table/header=true,
                        scale only axis
                    ]
                        \addplot[only marks, color=MidnightBlue, mark size=2pt, forget plot] table [x={x}, y expr={\thisrow{accuracy} * 100}]{\myTable}; 

                        \addplot[sharp plot, color=BurntOrange, very thick] table [x={xtest}, y expr={100 * \thisrow{ymean}}] {\myTable}; 

                        \addplot[name path=std_upper, draw=none] table [x={xtest}, y expr=100 * (\thisrow{ymean} + \stdScale * \thisrow{ystd})] {\myTable}; 
                        \addplot[name path=std_lower, draw=none] table [x={xtest}, y expr=100 * (\thisrow{ymean} - \stdScale * \thisrow{ystd})] {\myTable}; 
                        \addplot [fill=BurntOrange, opacity=0.25] fill between [of=std_upper and std_lower];

                        \pgfmathsetmacro{\newSlope}{100 * \slope}
                        \pgfmathsetmacro{\newIntercept}{100 * \intercept}
                        \addlegendentry{\(y = \pgfmathprintnumber{\newSlope} \cdot x
                        \pgfmathprintnumber[print sign]{\newIntercept}\)}
                    \end{axis}
                \end{tikzpicture}
                \caption{}
                \label{fig:task_entropy_miniImageNet}
            \end{subfigure}
            \hspace{2em}
            \begin{subfigure}[b]{0.2 \linewidth}
                \begin{tikzpicture}
                    \tikzstyle{every node}=[font=\small]
                    \pgfmathsetmacro{\slope}{-0.00085341}
                    \pgfmathsetmacro{\intercept}{0.8767791886648062}
                    \pgfmathsetmacro{\stdScale}{1}
                    \pgfmathsetmacro{\newSlope}{100 * \slope}
                    \pgfmathsetmacro{\newIntercept}{100 * \intercept}
                    
                    \pgfplotstableread[col sep=comma, header=true]{results/Task_distance_omniglot.csv} \myTable

                    \begin{axis}[
                        height = \linewidth,
                        width = \linewidth,
                        clip marker paths=true,
                        title = {Omniglot},
                        title style = {yshift=-0.5em},
                        enlarge y limits=true,
                        xlabel = {\(\mathrm{KL} \left[ \mathrm{test} \Vert \mathrm{train} \right]\)},
                        legend style={draw=none, anchor=north, at={(0.5, -0.3)}},
                        table/col sep=comma,
                        table/header=true,
                        scale only axis
                    ]
                        \addplot[only marks, color=MidnightBlue, mark size=2pt, forget plot] table [x={x}, y expr={\thisrow{accuracy} * 100}]{\myTable}; 

                        \addplot[sharp plot, color=BurntOrange, very thick] table [x={xtest}, y expr = {100 * \thisrow{ymean}}] {\myTable}; 

                        \addplot[name path=std_upper, draw=none] table [x={xtest}, y expr=100 * (\thisrow{ymean} + \stdScale * \thisrow{ystd})] {\myTable}; 
                        \addplot[name path=std_lower, draw=none] table [x={xtest}, y expr=100 * (\thisrow{ymean} - \stdScale * \thisrow{ystd})] {\myTable}; 
                        \addplot [fill=BurntOrange, opacity=0.25] fill between [of=std_upper and std_lower];

                        \pgfmathsetmacro{\newSlope}{100 * \slope}
                        \pgfmathsetmacro{\newIntercept}{100 * \intercept}
                        \addlegendentry{\(y = \pgfmathprintnumber{\newSlope} \cdot x
                        \pgfmathprintnumber[print sign]{\newIntercept}\)}
                    \end{axis}
                \end{tikzpicture}
                \caption{}
                \label{fig:test2train_omniglot_MAML}
            \end{subfigure}
            \hspace{2em}
            \begin{subfigure}[b]{0.2 \linewidth}
                \begin{tikzpicture}
                    \tikzstyle{every node}=[font=\small]
                    \pgfmathsetmacro{\slope}{-0.00050119}
                    \pgfmathsetmacro{\intercept}{0.9101944200456439}
                    \pgfmathsetmacro{\stdScale}{1}
                    \pgfmathsetmacro{\newSlope}{100 * \slope}
                    \pgfmathsetmacro{\newIntercept}{100 * \intercept}

                    \pgfplotstableread[col sep=comma, header=true]{results/Task_distance_miniImageNet_64.csv} \myTable

                    \begin{axis}[
                        height = \linewidth,
                        width = \linewidth,
                        clip marker paths=true,
                        title = {Mini-ImageNet},
                        title style = {yshift=-0.5em},
                        enlarge y limits=true,
                        xlabel = {\(\mathrm{KL} \left[ \mathrm{test} \Vert \mathrm{train} \right]\)},
                        legend style={draw=none, anchor=north, at={(0.5, -0.3)}},
                        table/col sep=comma,
                        table/header=true,
                        scale only axis
                    ]
                        \addplot[only marks, color=MidnightBlue, mark size=2pt, forget plot] table [x={x}, y expr={\thisrow{accuracy} * 100}]{\myTable}; 

                        \addplot[sharp plot, color=BurntOrange, very thick] table [x={xtest}, y expr = {100 * \thisrow{ymean}}] {\myTable}; 

                        \addplot[name path=std_upper, draw=none] table [x={xtest}, y expr=100 * (\thisrow{ymean} + \stdScale * \thisrow{ystd})] {\myTable}; 
                        \addplot[name path=std_lower, draw=none] table [x={xtest}, y expr=100 * (\thisrow{ymean} - \stdScale * \thisrow{ystd})] {\myTable}; 
                        \addplot [fill=BurntOrange, opacity=0.25] fill between [of=std_upper and std_lower];

                        \pgfmathsetmacro{\newSlope}{100 * \slope}
                        \pgfmathsetmacro{\newIntercept}{100 * \intercept}
                        \addlegendentry{\(y = \pgfmathprintnumber{\newSlope} \cdot x
                        \pgfmathprintnumber[print sign]{\newIntercept}\)}
                    \end{axis}
                \end{tikzpicture}
                \caption{}
                \label{fig:test2train_miniImageNet_MAML}
            \end{subfigure}
            \caption{Correlation diagrams between prediction accuracy made by MAML on 100 5-way 1-shot testing tasks versus: \protect\subref{fig:task_entropy_omniglot} and \subref{fig:task_entropy_miniImageNet} entropy of the inferred task-theme mixture distributions, and \subref{fig:test2train_omniglot_MAML} and \subref{fig:test2train_miniImageNet_MAML} the KL distances from testing to training tasks. The results show that largest the task entropy or distances, the worse the testing performance. The blue dots are the prediction made the MAML and PTM, the solid line is the mean of Bayesian Ridge regression, and the shaded areas correspond to \(\pm 1\) standard deviation around the mean.}
        \end{figure*}

    \subsection{Task distance matrix and correlation diagrams}
    \label{sec:correlation_diagram}
        Task distance matrix is used as one of the tools to qualitatively validate the prediction made by PTM. In particular, the hypothesis is that the PTM would predict small distances for tasks that are close together. Since the \say{labels} specifying the closedness of tasks are unknown, we utilise the hierarchical structure of Omniglot dataset to form tasks. Each task is generated by firstly sampling an alphabet, and then choosing characters in that alphabet. Under this strategy, tasks formed from the same alphabet would have small distances comparing to tasks from different alphabets. \figureautorefname~\ref{fig:task_distance_matrix} shows the task distances between 50 testing tasks of Omniglot dataset, where each block of 5 tasks on rows and columns of the task distance matrix corresponds to a group of tasks sampled from the same alphabet. The result, especially the square 5-task-by-5-task blocks along the diagonal, agrees well with the hypothesis. Note that the distance matrix shown in \figureautorefname~\ref{fig:task_distance_matrix} is asymmetric due to the asymmetric nature of the KL divergence used to measure task distance.

        We use a correlation diagram between prediction accuracy made by MAML and the \emph{task entropy} produced by PTM as another verification. Since the \emph{task entropy} denotes the uncertainty when modelling a task, we hypothesise that it proportionally relates to the difficulty when learning that task. To construct the correlation diagram, we firstly train a meta-learning model based on MAML using the training tasks of the two datasets, and evaluating the performance on 100 random testing tasks. Secondly, we calculate the \emph{task entropy} for those 100 testing tasks. Finally, we plot the prediction accuracy and task entropy in \figureautorefname{s~\ref{fig:task_entropy_omniglot}} and \ref{fig:task_entropy_miniImageNet}. The results on both datasets show that the higher the task uncertainty, the worse the test performance. This observation, therefore, agrees with our hypothesis about \emph{task entropy}.
        
        
        We conduct another correlation diagram between training-testing task distance and the test performance to verify further the proposed PTM. Our hypothesis is the inverse proportion between training-testing task distance and prediction accuracy. A similar experiment as in task uncertainty is carried out with a modification in which the task uncertainty is replaced by the average KL divergence between all training tasks to each testing task. Due to the extremely large number of training tasks, e.g. more than \(10^{12}\) unique 5-way tasks can be generated from both the two datasets, the calculation of the distance measure is infeasible. To make the training and testing tasks manageable, we randomly generate \(10,000\) tasks for training, and \(100\) tasks for testing. This results in \(1, 000, 000\) distances, which can be calculated in parallel with multiple computers. A testing task can be represented in the correlation diagram through its prediction accuracy and the average KL distance to training tasks, which is defined as:
        \begin{equation*}
            \overline{D}_{\mathrm{KL}}(\bm{\gamma}_{T + 1}) = \frac{1}{T} \sum_{i=1}^{T} D_{\mathrm{KL}}[q(\bm{\pi}; \bm{\gamma}_{T + 1}) \Vert q(\bm{\pi}; \bm{\gamma}_{i})].
        \end{equation*}
        The correlation diagrams for both datasets are then plotted in \figureautorefname{s~\ref{fig:test2train_omniglot_MAML}} and \ref{fig:test2train_miniImageNet_MAML}. The results agree well with our hypothesis, in which the further a testing task is from the training tasks, the worse the prediction accuracy. This enables us to use the new representation produced by PTM to analyse task similarity.
        
        \begin{figure*}[t]
            \centering
            \begin{subfigure}[b]{0.3 \linewidth}
                \centering
                \begin{tikzpicture}

                    \pgfplotstableread[col sep=comma, header=true]{results/Induction_curves.csv} \myTable

                    \begin{axis}[
                        height=0.8 \linewidth,
                        width=0.8 \linewidth,
                        xlabel={\textnumero~of training tasks (\(\times 1,000\))},
                        ylabel={Prediction accuracy (\%)},
                        ymin=40,
                        ymax=42.15,
                        table/col sep=comma,
                        table/header=true,
                        table/x=x,
                        scale only axis
                    ]
                        \addplot[mark=none, MidnightBlue, solid, very thick] table[y = {PTM}]{\myTable}; 
                        \addplot[mark=none, BurntOrange, dashdotted, very thick] table[y = {Task2Vec}]{\myTable}; 
                        \addplot[mark=none, PineGreen, dashed, very thick] table[y = {Worst case}]{\myTable}; 
                        \addplot[mark=none, BrickRed, dotted, very thick] table[y = {Random}]{\myTable}; 
                        \end{axis}
                \end{tikzpicture}
                \caption{Induction}
                \label{fig:accuracy_induction}
            \end{subfigure}
            \hfill
            \begin{subfigure}[b]{0.3 \linewidth}
                \centering
                \begin{tikzpicture}

                    \pgfplotstableread[col sep=comma, header=true]{results/Transduction_curves.csv} \myTable

                    \begin{axis}[
                        height=0.8 \linewidth,
                        width=0.8 \linewidth,
                        xlabel={\textnumero~of training tasks (\(\times 1,000\))},
                        ylabel={\vphantom{h}},
                        ymin=40,
                        ymax=42.15,
                        restrict x to domain=0:300,
                        table/col sep=comma,
                        table/x=x,
                        table/header=true,
                        table/skip rows between index={602}{707},
                        scale only axis,
                        legend entries={PTM, Task2Vec, Worst case, Random},
                        legend style={draw=none, font=\footnotesize},
                        legend cell align={left},
                        legend pos=south east
                    ]
                        \addplot[mark=none, MidnightBlue, solid, very thick] table[y={PTM}]{\myTable}; 
                        \addplot[mark=none, BurntOrange, dashdotted, very thick] table[y= {Task2Vec}]{\myTable}; 
                        \addplot[mark=none, PineGreen, dashed, very thick] table[y = {Worst case}]{\myTable}; 
                        \addplot[mark=none, BrickRed, dotted, very thick] table[y = {Random}]{\myTable}; 
                        \end{axis}
                \end{tikzpicture}
                \caption{Transduction}
                \label{fig:accuracy_transduction}
            \end{subfigure}
            \hfill
            \begin{subfigure}[b]{0.3 \linewidth}
                \centering
                \begin{tikzpicture}
                    \begin{axis}[
                        height=0.8 \linewidth,
                        width=0.8 \linewidth,
                        ybar=0pt,
                        ymin=39.5,
                        ymax=42.5,
                        legend style={draw=none, font=\footnotesize, /tikz/every even column/.append style={column sep=0.5em}, anchor=north, legend columns=2},
                        legend pos=north west,
                        legend image post style={scale=0.75},
                        legend cell align={left},
                        ylabel={\vphantom{h}},
                        symbolic x coords={MAML, Protonet, ABML},
                        xtick=data,
                        xticklabel style = {font=\small},
                        xlabel={\vphantom{Hello}},
                        scale only axis,
                        enlarge x limits=0.2,
                        bar width = 8pt
                    ]
                        \addplot[style={fill=MidnightBlue, draw=none}, error bars/.cd, y dir=both, y explicit] coordinates {
                            (MAML, 41.36) +- (0, 0.14)
                            (Protonet, 41.12) +- (0, 0.14)
                            (ABML, 41.59) +- (0, 0.13)
                        };
                        \addplot[style={fill=BurntOrange, draw=none}, error bars/.cd, y dir=both, y explicit] coordinates {
                            (MAML, 41.23) +- (0, 0.14)
                            (Protonet, 40.75) +- (0, 0.14)
                            (ABML, 41.33) +- (0, 0.14)
                        };
                        \addplot[style={fill=PineGreen, draw=none}, error bars/.cd, y dir=both, y explicit] coordinates {
                            (MAML, 40.93) +- (0, 0.14)
                            (Protonet, 40.25) +- (0, 0.14)
                            (ABML, 41.09) +- (0, 0.14)
                        };
                        \addplot[style={fill=BrickRed, draw=none}, error bars/.cd, y dir=both, y explicit] coordinates {
                            (MAML, 40.44) +- (0, 0.14)
                            (Protonet, 39.64) +- (0, 0.14)
                            (ABML, 40.48) +- (0, 0.14)
                        };
                        \legend{PTM, Task2Vec, Worst case, Random}
                    \end{axis}
                \end{tikzpicture}
                \caption{Final test results}
                \label{fig:accuracy_induction_test}
            \end{subfigure}
            \caption{Exponential weighted moving average (EWMA) of prediction accuracy made by MAML following the lifelong learning for 100 random 5-way 1-shot tasks sampled from mini-ImageNet testing set: \protect\subref{fig:accuracy_induction} inductive setting, and \subref{fig:accuracy_transduction} transductive setting. The EWMA weight is set to 0.98 to smooth the noisy signal. \subref{fig:accuracy_induction_test} Prediction accuracy made by models trained on different task selection approaches on all 5-way 1-shot testing tasks generated from mini-ImageNet. The error bars correspond to 95 percent confident interval.}
            \label{fig:accuracy_lifelong_learning}
        \end{figure*}
        
    \subsection{Lifelong few-shot meta-learning}
    \label{sec:lifelong_learning}
        To further evaluate PTM, we conduct experiments following the lifelong learning framework~\citep{ruvolo2013active} with slight modification where the supervised tasks are replaced by 5-way 1-shot learning episodes. More precisely, the setting consists of a meta-learning model and a pool of \(T_{\mathrm{pool}}\) tasks. At each time step, a task selected from the pool is used to update the meta-learning model, and discarded from the pool. A new task is then added to the pool to maintain \(T_{\mathrm{pool}}\)  tasks available for learning. The criterion for selecting a task to update the meta-learning model will depend on the objective of interest. Two common objectives often observed in practice are:
        \begin{itemize}
            \item \textbf{Induction:} the selected training task is expected to encourage the meta-learning model to be able to rapidly adapt to \textbf{any} future task,
            \item \textbf{Transduction:} the selected training task is targeted toward one or many \textbf{specific} testing tasks.
        \end{itemize}

        In the induction setting, the performance of the meta-learning model trained on tasks selected by PTM is compared with three baselines: Task2Vec~\citep{achille2019task2vec}, the \say{worst-case} approach~\citep{collins2020task} and random selection. 
        For the PTM, the selection criteria is based on the task entropy specified in \cref{sec:methodology}, where the training task with highest entropy is chosen for the learning. 
        For Task2Vec, tasks with large embedding norm are reported as difficult to learn. Hence, we pick the one with the largest L1 norm produced by Task2Vec as the training task.
        Originally, Task2Vec requires fine-tuning a pre-trained network (known as probe network) on labelled data of a task. This fine-tuning step is, however, infeasible for few-shot learning due to the insufficient number of labelled data. We address this issue by training a MAML-based network to use as a probe network. When given few-shot data of a training task, the MAML-based probe network perform gradient update to adapt to that task. The task-specific embedding can, therefore, be calculated using the adapted probe network. We follow the Monte Carlo approach specified in the public code of Task2Vec to calculate the corresponding task embedding.
        For the \say{worst-case} approach, the training task that results in the highest loss for the current meta-learning model is selected. Due to this nature, the \say{worst-case} approach requires to evaluate all losses for each task in the pool at every time step, leading to an extensive computation and might not scale well when the number of tasks in the pool is large. For simplicity, we use MAML to train the meta-learning model of interest for each selection strategy.

        The transduction setting follows a similar setup as the induction case, but the testing tasks, including the labelled and unlabelled data, are known during training. For PTM, the average KL distances between all testing tasks to each training task in the task pool are calculated, and the training task with smallest average distance is selected. For Task2Vec, the proposed cosine distance between normalised task embeddings is used to calculate the average distance between all testing tasks to each training task~\citep{achille2019task2vec}. Similar to PTM, the training task with the smallest distance is prioritised for the learning. For the \say{worst-case} approach, the entropy of the prediction \(\hat{y}\) on \(C\)-way testing tasks is used as the measure:
        \begin{equation*}
            S_{T + 1} = -\sum_{c=1}^{C} \hat{y}_{c} \ln \hat{y}_{c},
        \end{equation*}
        and the task that contributed to the highest entropy at prediction is chosen~\citep{mackay1992evidence}. The \say{worst-case} approach, therefore, requires \(T_{\mathrm{pool}}\) trials at every time step. In each trial, the current meta-model is adapted to each training task in the pool, and then the average prediction entropy on all testing tasks is calculated. This results in an extremely extensive computation.
        
        Four MAML-based meta-learning models are initialised identically and trained on the tasks selected from a pool of \(T_{\mathrm{pool}} = 200\) tasks according to the four criteria mentioned above. \figureautorefname{s~\ref{fig:accuracy_induction}} and \ref{fig:accuracy_transduction} show the testing results on 100 random mini-ImageNet tasks after every 500 time steps. Note that the plotted results are smoothed by the exponential weighted moving average with a weight of 0.98 to ease the visualisation. In general, PTM, Task2Vec and \say{worst-case} can generalise better than random task selection. In addition, the model trained with tasks chosen by PTM performs slightly better than Task2Vec and the \say{worst-case} approach in both settings. This observation might be explained based on the designated purpose of Task2Vec and the \say{worst-case} approach. Task2Vec requires a sufficient number of labelled data to fine-tune its probe network to calculate task embedding. Hence, it might not work well in few-shot learning. For the \say{worst-case}, tasks are selected according to a measure based on the current meta-model without taking task relatedness into account. PTM, however, has a weakness in active selection since the approach only focuses on task uncertainty or task similarity without considering the current state of the meta-learning model. Nevertheless, PTM still provides a good selection criterion comparing to Task2Vec and the \say{worst-case} approaches. Note that although the active task selection is able to select the best task within the pool, there might be the case where all remaining tasks in the pool are uninformative, resulting in overfitting as observed in \figureautorefname~\ref{fig:accuracy_induction}. However, for simplicity, no additional mechanism is integrated to decide whether to learn from the selected task, or simply discarded from the pool. We believe that adding L2 regularisation or applying early stopping based on a validation set of tasks will help with this overfitting issue.
        
        To further compare, we implement two additional meta-learning algorithms: Prototypical Networks~\citep{snell2017prototypical} and Amortised Bayesian meta-learning (ABML)~\citep{ravi2019amortized} and show results for the induction setting on all available testing 5-way 1-shot tasks of mini-ImageNet in \figureautorefname{~\ref{fig:accuracy_induction_test}}. Again, the prediction accuracy made by the model trained on tasks selected by PTM outperforms other baselines, especially the random one by a large margin.

%% file: Conclusion.tex
\section{Conclusion}
\label{sec:conclusion}
    We propose a generative approach based on variational auto-encoding and LDA adopted in topic modelling to model tasks used in meta-learning. Under this modelling approach, the dataset associated with a task can be expressed as a mixture model of finite Gaussian distributions, where each task differs at the mixture coefficients. An online VI method is presented to infer the parameters of the Gaussian task-theme distributions. The obtained model allows us to represent a task by its variational distribution of mixture coefficient in a latent task-theme simplex, enabling the quantification of either the task uncertainty or task similarity for active task selection.

%% file: Appendices.tex
\appendix
    \section{Calculation of each term in the ELBO}
    \label{sec:elbo_calculation}
        As described in \sectionautorefname~\ref{sec:methodology}, the variational distributions for \(\mathbf{u}, \mathbf{z}\) and \(\bm{\pi}\) are:
        \begin{align}
            q(\mathbf{u}_{in}; \phi) & = \mathcal{N} \left(\mathbf{u}_{in}; \mathbf{m}_{in}, \left(\mathbf{s}_{in}\right)^{2} \mathbf{I} \right) \tag{\ref{eq:q_u}}\\
            q(\bm{\pi}_{i}; \bm{\gamma}_{i}) & = \mathrm{Dirichlet}\left( \bm{\pi}_{i}; \bm{\gamma}_{i} \right) \tag{\ref{eq:q_pi}}\\
    	    q(\mathbf{z}_{in}; \mathbf{r}_{in}) & = \mathrm{Categorical} \left( \mathbf{z}_{in}; \mathbf{r}_{in} \right) \tag{\ref{eq:q_z}}.
        \end{align}

        \subsection{\texorpdfstring{\( \mathbb{E}_{q(\mathbf{u}_{i}; \bm{\mu}_{u_{i}}, \bm{\Sigma}_{u_{i}})} \mathbb{E}_{q(\mathbf{z}_{i}, \bm{\pi}_{i})} \left[ \ln p(\mathbf{u}_{i} | \mathbf{z}_{i}, \bm{\mu}, \bm{\Sigma}) \right] \)}{}}
            \begin{equation}
                \begin{aligned}[b]
                    \mathbb{E}_{q(\mathbf{z}_{i}, \bm{\pi}_{i})} \left[ \ln p(\mathbf{u}_{i} | \mathbf{z}_{i}, \bm{\mu}, \bm{\Sigma}) \right] & = \sum_{n=1}^{N} \sum_{k=1}^{K} r_{ink} \ln p(\mathbf{u}_{in} | \bm{\mu}_{k}, \bm{\Sigma}_{k}) \\
                    & = \sum_{n=1}^{N} \sum_{k=1}^{K} r_{ink} \ln \mathcal{N} (\mathbf{u}_{in} | \bm{\mu}_{k}, \bm{\Sigma}_{k}).
                \end{aligned}
            \end{equation}
            Hence:
            \begin{equation}
                \begin{aligned}[b]
                \mathbb{E}_{q(\mathbf{u}_{i}; \bm{\mu}_{u_{i}}, \bm{\Sigma}_{u_{i}})} \mathbb{E}_{q(\mathbf{z}_{i}, \bm{\pi}_{i})} \left[ \ln p(\mathbf{u}_{i} | \mathbf{z}_{i}, \bm{\mu}, \bm{\Sigma}) \right] & = \sum_{n=1}^{N} \sum_{k=1}^{K} r_{ink} \underbrace{\mathbb{E}_{q(\mathbf{u}_{i}; \bm{\mu}_{u_{i}}, \bm{\Sigma}_{u_{i}})} \left[ \ln \mathcal{N} (\mathbf{u}_{i} | \bm{\mu}_{k}, \bm{\Sigma}_{k}) \right]}_{\text{cross-entropy between 2 Gaussians}} \\
                & = \sum_{n=1}^{N} \sum_{k=1}^{K} r_{ink} \left[ - \frac{1}{2} \mathrm{tr}(\bm{\Sigma}_{k}^{-1} \bm{\Sigma}_{u_{in}}) + \ln \mathcal{N}(\bm{\mu}_{u_{in}}; \bm{\mu}_{k}, \bm{\Sigma}_{k}) \right].
                \end{aligned}
            \end{equation}
            
        \subsection{\texorpdfstring{\( \mathbb{E}_{q(\mathbf{u}_{i}; \bm{\mu}_{u_{i}}, \bm{\Sigma}_{u_{i}})} \mathbb{E}_{q(\mathbf{z}_{i}, \bm{\pi}_{i})} \left[ \ln p(\mathbf{z}_{i} | \bm{\pi}_{i}) \right] \)}{}}
            \begin{equation}
                \begin{aligned}[b]
                    \mathbb{E}_{q(\mathbf{u}_{i}; \bm{\mu}_{u_{i}}, \bm{\Sigma}_{u_{i}})} \mathbb{E}_{q(\mathbf{z}_{i}, \bm{\pi}_{i})} \left[ \ln p(\mathbf{z}_{i} | \bm{\pi}_{i}) \right] & = \sum_{n=1}^{N} \sum_{k=1}^{K} r_{ink} \int \mathrm{Dir}_{K} (\bm{\pi}_{i}; \bm{\gamma}_{i}) \ln \pi_{ik} \dd{\pi_{ik}} \\
                    & = \sum_{n=1}^{N} \sum_{k=1}^{K} r_{ink} \ln \Tilde{\pi}_{ik},
                \end{aligned}
            \end{equation}
            where:
            \begin{equation}
            \boxed{
                \ln \Tilde{\pi}_{ik} = \psi(\gamma_{ik}) - \psi\left( \sum_{j=1}^{K} \gamma_{ij} \right).
            }
            \label{eq:log_pi_tilde}
            \end{equation}
    
        \subsection{\texorpdfstring{\( \mathbb{E}_{q(\mathbf{u}_{i}; \bm{\mu}_{u_{i}}, \bm{\Sigma}_{u_{i}})} \mathbb{E}_{q(\mathbf{z}_{i}, \bm{\pi}_{i})} \left[ \ln p(\bm{\pi}_{i} | \bm{\alpha}) \right] \)}{}}
            \begin{equation}
                \begin{aligned}[b]
                    \mathbb{E}_{q(\mathbf{u}_{i}; \bm{\mu}_{u_{i}}, \bm{\Sigma}_{u_{i}})} \mathbb{E}_{q(\mathbf{z}_{i}, \bm{\pi}_{i})} \left[ \ln p(\bm{\pi}_{i} | \bm{\alpha}) \right] & = \mathbb{E}_{\mathrm{Dir}(\bm{\pi}_{i}; \bm{\gamma}_{i})} \left[ \ln \Gamma \left( \sum_{j=1}^{K} \alpha_{j} \right) - \left[\sum_{k=1}^{K} \ln \Gamma(\alpha_{k}) - (\alpha_{k} - 1) \ln \pi_{ik} \right] \right] \\
                    & = \left[\ln \Gamma \left(\sum_{j=1}^{K} \alpha_{j} \right) - \sum_{k=1}^{K} \ln\Gamma(\alpha_{k}) \right] + \sum_{k=1}^{K} (\alpha_{k} - 1) \ln\Tilde{\pi}_{ik}.
                \end{aligned}
            \end{equation}
    
        \subsection{\texorpdfstring{\( \mathbb{E}_{q(\mathbf{u}_{i}; \bm{\mu}_{u_{i}}, \bm{\Sigma}_{u_{i}})} \mathbb{E}_{q(\mathbf{z}_{i}, \bm{\pi}_{i})} \left[ \ln q(\mathbf{z}_{i} | \mathbf{r}_{i}) \right] \)}{}}
            \begin{equation}
                \mathbb{E}_{q(\mathbf{u}_{i}; \bm{\mu}_{u_{i}}, \bm{\Sigma}_{u_{i}})} \mathbb{E}_{q(\mathbf{z}_{i}, \bm{\pi}_{i})} \left[ \ln q(\mathbf{z}_{i} | \mathbf{r}_{i}) \right] = \sum_{n=1}^{N} \sum_{k=1}^{K} r_{ink} \ln r_{ink}.
            \end{equation}
        
        \subsection{\texorpdfstring{\( \mathbb{E}_{q(\mathbf{u}_{i}; \bm{\mu}_{u_{i}}, \bm{\Sigma}_{u_{i}})} \mathbb{E}_{q(\mathbf{z}_{i}, \bm{\pi}_{i})} \left[ \ln q(\bm{\pi}_{i} | \bm{\gamma}_{i}) \right] \)}{}}
            \begin{equation}
                \mathbb{E}_{q(\mathbf{u}_{i}; \bm{\mu}_{u_{i}}, \bm{\Sigma}_{u_{i}})} \mathbb{E}_{q(\mathbf{z}_{i}, \bm{\pi}_{i})} \left[ \ln q(\bm{\pi}_{i} | \bm{\gamma}_{i}) \right] = \ln\Gamma \left( \sum_{j=1}^{K} \gamma_{ij} \right) - \sum_{k=1}^{K} \left[ \ln\Gamma(\gamma_{ik}) - (\gamma_{ik} - 1) \ln \Tilde{\pi}_{ik} \right].
            \end{equation}
    
    \section{Maximisation of the ELBO}
    \label{sec:elbo_optimisation}
        Since the ELBO can be evaluated as shown in \appendixname~\ref{sec:elbo_calculation}, we can maximise the ELBO w.r.t. \say{task-specific} variational parameters by taking derivative, setting it to zero and solving for the parameters of interest.
            
        \subsection{Variational categorical distribution}
        \label{sec:vi_z}
            Note that:
            \begin{equation}
                \sum_{k=1}^{K} r_{ink} = 1.
            \end{equation}
            The derivative of \(\mathsf{L}_{i}\) with respect to \(r_{ink}\) can be expressed as:
            \begin{equation}
                \begin{aligned}[b]
                \frac{\partial\mathsf{L}}{\partial r_{ink}} = - \frac{1}{2} \mathrm{tr}(\bm{\Sigma}_{k}^{-1} \bm{\Sigma}_{u_{in}}) + \ln \mathcal{N}(\bm{\mu}_{u_{in}}; \bm{\mu}_{k}, \bm{\Sigma}_{k}) + \ln \Tilde{\pi}_{ik} - \ln r_{ink} - 1 + \lambda,
                \end{aligned}
            \end{equation}
            where: \(\lambda\) is the Lagrange multiplier and \(\ln \Tilde{\pi}_{ik}\) is defined in Eq.~\eqref{eq:log_pi_tilde}.
            Setting the derivative to zero and solving for \(r_{ink}\) give:
            \begin{equation}
            \boxed{
                r_{ink} \propto \exp \left[ - \frac{1}{2} \mathrm{tr}(\bm{\Sigma}_{k}^{-1} \bm{\Sigma}_{u_{in}}) + \ln \mathcal{N}(\bm{\mu}_{u_{in}}; \bm{\mu}_{k}, \bm{\Sigma}_{k}) + \ln \Tilde{\pi}_{ik} \right].
            }
            \label{eq:r}
            \end{equation}
            
        \subsection{Variational Dirichlet distribution}
        \label{sec:vi_pi}
            The lower-bound related to \(\gamma_{ik}\) can be written as:
            \begin{equation}
                \begin{aligned}[b]
                    \mathsf{L} & = \sum_{k=1}^{K} \sum_{n=1}^{N} r_{ink} \ln \Tilde{\pi}_{ik} + \sum_{k=1}^{K} (\alpha_{k} - 1) \ln\Tilde{\pi}_{ik} - \ln\Gamma \left( \sum_{j=1}^{K} \gamma_{ij} \right) + \sum_{k=1}^{K} \left[ \ln\Gamma(\gamma_{ik}) - (\gamma_{ik} - 1) \ln \Tilde{\pi}_{ik} \right]\\
                    & = -\ln\Gamma \left( \sum_{j=1}^{K} \gamma_{ij} \right) + \sum_{k=1}^{K} \ln \Tilde{\pi}_{ik} \left( \alpha_{k} - \gamma_{ik} + \sum_{n=1}^{N} r_{ink} \right) + \ln\Gamma(\gamma_{ik}) \\
                    & = -\ln\Gamma \left( \sum_{j=1}^{K} \gamma_{ij} \right) + \sum_{k=1}^{K} \left[ \psi(\gamma_{ik}) - \psi\left( \sum_{j=1}^{K} \gamma_{ij} \right) \right] \left( \alpha_{k} - \gamma_{ik} + \sum_{n=1}^{N} r_{ink} \right) + \ln\Gamma(\gamma_{ik}).
                \end{aligned}
            \end{equation}
            
            Hence, the lower-bound related to \(\gamma_{ik}\) is:
            \begin{equation}
                \begin{aligned}[b]
                    \mathsf{L}[\gamma_{ik}] & = -\ln\Gamma \left( \sum_{j=1}^{K} \gamma_{ij} \right) + \psi(\gamma_{ik}) \left( \alpha_{k} - \gamma_{ik} + \sum_{n=1}^{N} r_{ink} \right) \\
                    & \qquad - \psi\left( \sum_{j=1}^{K} \gamma_{ij} \right) \left( \sum_{j=1}^{K} \alpha_{j} - \gamma_{ij} + \sum_{n=1}^{N} r_{inj} \right)  + \ln\Gamma(\gamma_{ik})
                \end{aligned}
            \end{equation}
            
            Taking derivative w.r.t. \(\gamma_{ik}\) gives:
            \begin{equation}
                \begin{aligned}[b]
                    \frac{\partial \mathsf{L}}{\partial \gamma_{ik}} & = -\psi \left( \sum_{j=1}^{K} \gamma_{ij} \right) + \Psi(\gamma_{ik}) \left( \alpha_{k} - \gamma_{ik} + \sum_{n=1}^{N} r_{ink} \right) - \psi(\gamma_{ik}) \\
                    & \quad - \Psi \left( \sum_{j=1}^{K} \right) \left( \sum_{j=1}^{K} \alpha_{j} - \gamma_{ij} + \sum_{n=1}^{N} r_{inj} \right) + \psi \left( \sum_{j=1}^{K} \gamma_{ij} \right) + \psi(\gamma_{ik}) \\
                    & = \Psi(\gamma_{ik}) \left( \alpha_{k} - \gamma_{ik} + \sum_{n=1}^{N} r_{ink} \right) - \Psi \left( \sum_{j=1}^{K} \right) \sum_{j=1}^{K} \alpha_{j} - \gamma_{ij} + \sum_{n=1}^{N} r_{inj},
                \end{aligned}
            \end{equation}
            where \(\Psi(.)\) is the trigamma function.
            
            Setting the derivative to zero yields a maximum at:
            \begin{equation}
                \boxed{
                    \gamma_{ik} = \alpha_{k} + N_{ik},
                }
                \label{eq:gamma}
            \end{equation}
            where:
            \begin{equation}
                N_{ik} = \sum_{n=1}^{N}r_{ink}.
            \end{equation}
            
        \subsection{Maximum likelihood for the task-theme \texorpdfstring{\(\bm{\mu}_{k}\) and \(\bm{\Sigma}_{k}\)}{}}
            The terms in the objective function relating to \(\mu_{k}\) can be written as:
            \begin{equation}
                \begin{aligned}[b]
                    \mathsf{L}[\mu_{k}] & = \sum_{i=1}^{T} \sum_{n=1}^{N} r_{ink} \ln \mathcal{N} \left( \bm{\mu}_{u_{in}}; \bm{\mu}_{k}, \bm{\Sigma}_{k} \right)
                    \\
                    & = -\frac{1}{2} \sum_{i=1}^{T} \sum_{n=1}^{N} r_{ink} \left( \bm{\mu}_{u_{in}} - \bm{\mu}_{k} \right)^{\top} \bm{\Sigma}_{k}^{-1} \left( \bm{\mu}_{u_{in}} - \bm{\mu}_{k} \right)
                \end{aligned}
            \end{equation}
            
            Taking derivative w.r.t. \(\bm{\mu}_{k}\) gives:
            \begin{equation}
                \pdv{\mathsf{L}}{\bm{\mu}_{k}} = \sum_{i=1}^{T} \sum_{n=1}^{N} r_{ink} \bm{\Sigma}_{k}^{-1} (\bm{\mu}_{u_{in}} - \bm{\mu}_{k}).
            \end{equation}
            
            Setting the derivative to zero yields a maximum at:
            \begin{equation}
                \boxed{
                    \bm{\mu}_{k} = \frac{\sum_{i=1}^{T} \sum_{n=1}^{N} r_{ink} \bm{\mu}_{u_{in}}}{\sum_{i=1}^{T} N_{ik}}.
                }
                \label{eq:local_mean}
            \end{equation}
            
            The terms in the objective function relating to \(\bm{\Sigma}_{k}\) is given as:
            \begin{equation}
                \begin{aligned}[b]
                    \mathsf{L} & = \sum_{i=1}^{T} \sum_{n=1}^{N} r_{ink} \left[ - \frac{1}{2} \mathrm{tr}(\bm{\Sigma}_{k}^{-1} \bm{\Sigma}_{u_{in}}) + \ln \mathcal{N}(\bm{\mu}_{u_{in}}; \bm{\mu}_{k}, \bm{\Sigma}_{k}) \right] \\
                    & = -\frac{1}{2} \sum_{i=1}^{T} \sum_{n=1}^{N} r_{ink} \left[ \mathrm{tr}(\bm{\Sigma}_{k}^{-1} \bm{\Sigma}_{u_{in}}) + d \ln (2\pi) + \ln |\bm{\Sigma}_{k}| + \left( \bm{\mu}_{u_{in}} - \bm{\mu}_{k} \right)^{\top} \bm{\Sigma}_{k}^{-1} \left( \bm{\mu}_{u_{in}} - \bm{\mu}_{k} \right) \right].
                \end{aligned}
            \end{equation}
            
            Taking derivative w.r.t. \(\bm{\Sigma}_{k}\) gives:
            \begin{equation}
                \begin{aligned}
                    \pdv{\mathsf{L}}{\bm{\Sigma}_{k}} & = -\frac{1}{2} \sum_{i=1}^{T} \sum_{n=1}^{N} r_{ink} \left[ - \bm{\Sigma}_{k}^{-1} \bm{\Sigma}_{u_{in}} \bm{\Sigma}_{k}^{-1} + \bm{\Sigma}_{k}^{-1} - \bm{\Sigma}_{k}^{-1} \left(\bm{\mu}_{u_{in}} - \bm{\mu}_{k} \right) \left(\bm{\mu}_{u_{in}} - \bm{\mu}_{k} \right)^{\top} \bm{\Sigma}_{k}^{-1} \right]\\
                    & = \frac{1}{2} \sum_{i=1}^{T} \sum_{n=1}^{N} r_{ink} \left[ \bm{\Sigma}_{k}^{-1} \bm{\Sigma}_{u_{in}} - \mathbf{I} + \bm{\Sigma}_{k}^{-1} \left(\bm{\mu}_{u_{in}} - \bm{\mu}_{k} \right) \left(\bm{\mu}_{u_{in}} - \bm{\mu}_{k} \right)^{\top} \right] \bm{\Sigma}_{k}^{-1}.
                \end{aligned}
            \end{equation}
            
            Setting the derivative to zero gives:
            \begin{equation}
            \boxed{
                \begin{aligned}
                    \bm{\Sigma}_{k} & = \frac{1}{\sum_{i=1}^{T} N_{ik}} \sum_{i=1}^{T} \sum_{n=1}^{N} r_{ink} \left[ \bm{\Sigma}_{u_{in}} + \left(\bm{\mu}_{u_{in}} - \bm{\mu}_{k} \right) \left(\bm{\mu}_{u_{in}} - \bm{\mu}_{k} \right)^{\top} \right].
                \end{aligned}
            }
            \label{eq:local_cov}
            \end{equation}
        
        \subsection{Maximum likelihood for \texorpdfstring{\(\bm{\alpha}\)}{alpha}}
            The lower-bound with terms relating to \(\alpha_{k}\) can be expressed as:
            \begin{equation}
                \mathsf{L} = T \left[\ln \Gamma \left(\sum_{j=1}^{K} \alpha_{j} \right) - \sum_{k=1}^{K} \ln\Gamma(\alpha_{k}) \right] + \sum_{i=1}^{T} \sum_{k=1}^{K} (\alpha_{k} - 1) \left[ \psi(\gamma_{ik}) - \psi\left( \sum_{j=1}^{K} \gamma_{ij} \right) \right].
            \end{equation}
            
            Taking derivative w.r.t. \(\alpha_{k}\) gives:
            \begin{equation}
                \begin{aligned}
                g_{k} = \pdv{\mathsf{L}}{\alpha_{k}} & = T \left[ \psi \left(\sum_{j=1}^{K} \alpha_{j} \right) - \psi (\alpha_{k}) \right] + \sum_{i=1}^{T} \left[ \psi(\gamma_{ik}) - \psi\left( \sum_{j=1}^{K} \gamma_{ij} \right) \right].
                \end{aligned}
            \end{equation}
            
            The second derivative is, therefore, obtained as:
            \begin{equation}
                \begin{aligned}[b]
                    \frac{\partial^{2} \mathsf{L}}{\partial \alpha_{k} \partial \alpha_{k^{\prime}}} & = T \left[ \Psi \left(\sum_{j=1}^{K} \alpha_{j} \right) - \delta(k - k^{\prime}) \Psi (\alpha_{k}) \right].
                \end{aligned}
            \end{equation}
            The Hessian can be written in matrix form~\citep{minka2000estimating} as:
            \begin{align}
                \mathbf{H} & = \mathbf{Q} + \mathbf{11}^{T} a \\
                q_{kk^{\prime}} & = - T \delta(k - k^{\prime}) \Psi (\alpha_{k}) \\
                a & = T  \Psi \left(\sum_{j=1}^{K} \alpha_{j} \right).
            \end{align}
            One Newton step is therefore:
            \begin{align}
                \bm{\alpha} & \gets \bm{\alpha} - \mathbf{H}^{-1} \mathbf{g} \\
                (\mathbf{H}^{-1} \mathbf{g})_{k} & = \frac{g_{k} - b}{q_{kk}},
                \label{eq:inverse_hessian_gradient}
            \end{align}
            where:
            \begin{equation}
                b = \frac{\sum_{j=1}^{K} g_{j} / q_{jj}}{1 / a + \sum_{j=1}^{K} 1/ q_{jj}}.
            \end{equation}